\documentclass{article}
\usepackage{spconf,amsmath,amssymb,graphicx,booktabs,array}

\newcommand{\ba}{\text{BA}}

\title{Byzantine-Robust Federated Fire Detection with a Rotating Coordinator}

\name{Georgia Argyrou$^{\star\dagger}$ \qquad Aymen Bahroun$^{\dagger}$ \qquad Hedi Fendri$^{\dagger}$ \qquad Alexander Jung$^{\star}$
\thanks{This paper presents results from the first author's M.Sc.\ thesis at
Aalto University~\cite{argyrou2026}, carried out in collaboration with Kudelski
Labs. Correspondence: Georgia Argyrou, \texttt{argeorgiaa@gmail.com};
Alexander Jung, \texttt{alex.jung@aalto.fi}.}}
\address{$^{\star}$Department of Computer Science, Aalto University, Espoo, Finland \\
         $^{\dagger}$Kudelski Labs}

\begin{document}
\ninept
\maketitle

\begin{abstract}
We study the application of federated learning (FL) to indoor fire detection.
Such fire-detection systems use edge cameras that record sensitive footage
which cannot easily be collected at a central server. Existing federated
solutions leave three practical obstacles unaddressed: limited uplink
bandwidth, Byzantine (malicious or faulty) clients, and unconditional trust in
a single, permanently fixed aggregation server. Our main contributions address
all three. In particular, we provide (i) a curated indoor fire-detection
dataset assembled from eight public sources; (ii) an edge-deployable detector
whose model updates are compressed up to $10\times$ with only a small loss in
balanced accuracy; and (iii) a semi-decentralized Byzantine-robust FL method
that combines history-aware aggregation with a rotating coordinator, evicting
stealthy attacks that per-round filters miss while removing the fixed-server
single point of failure. On the held-out test set the rotating-coordinator
method matches its fixed-server counterpart in accuracy and detection speed,
and a physically distributed six-node cloud deployment confirms feasibility.
\end{abstract}

\begin{keywords}
Federated learning, Byzantine robustness, communication efficiency, fire
detection, edge AI\@.
\end{keywords}

\section{Introduction}\label{sec:intro}
Indoor fires spread rapidly in confined, cluttered spaces and leave little time
for evacuation~\cite{drysdale2011}. Conventional smoke and heat detectors react only after a fire
has produced enough particulate or heat to cross a threshold and are prone to
false alarms~\cite{fonollosa2018}. Cameras already installed in many buildings offer an
alternative: convolutional neural networks can
detect the subtle visual cues of an incipient fire well before a physical sensor
activates~\cite{muhammad2018}.

Realizing this at scale runs into a fundamental obstacle. The cameras best
positioned to supply training data are spread across many buildings and
organizations under different ownership and privacy requirements, and
continuously streaming high-resolution footage off constrained devices is costly.
Federated learning (FL)~\cite{mcmahan2017} keeps raw data on-device and exchanges
only model updates, making it a natural fit for privacy-sensitive fire
surveillance. A federated setting, however,
introduces challenges absent from the centralized formulation. First, client
data are statistically heterogeneous---not independent and identically
distributed (non-IID). Data heterogeneity also challenges centralized ML
systems, but in FL each client updates the model using only its own local
data, which can additionally slow convergence.
Second, a single compromised edge camera can submit an arbitrarily
corrupted update---a \emph{Byzantine} attack~\cite{blanchard2017}---and hijack the
global model. Third, transmitting full updates every round is expensive for
bandwidth-limited cameras~\cite{konecny2016}. Finally, server-based FL places
unconditional trust in one permanently fixed coordinator, a single point of
failure that is especially undesirable in a safety-critical application.

\noindent\textbf{Related work.}
IOFireNet~\cite{panneerselvam2024} federates local MobileNet fire detectors but
does not address adversarial clients or uplink compression. Our method is
robust against Byzantine clients and compresses the uplink up to $10\times$.
FireFedXNet~\cite{hussein2025} trains residential MobileNet-V2 clients with
model-weight averaging and Grad-CAM++ explanations, but does not consider
Byzantine clients or coordinator failure; we address both and target
heterogeneous non-residential indoor scenes. FedMA~\cite{salam2025}
reaches high accuracy with a bio-inspired optimizer but at a high per-round
training cost, whereas our frozen-backbone design keeps rounds lightweight.
Finally, the rotating-server approaches in~\cite{wang2025,chang2023} remove the
fixed coordinator but do not address carrying a history-aware Byzantine detector's
per-client state across handoffs; ours is, to our knowledge, the first rotation
to preserve that state, which SafeguardSGD requires. No prior work offers a
heterogeneous indoor dataset (C1), a compressed edge-deployable federated
detector (C2), and Byzantine-robust aggregation on a fixed-server-free rotating
coordinator (C3) together---the gaps we close. We make
three contributions:
\begin{itemize}
  \item \textbf{(C1) Dataset.} A curated indoor fire-detection dataset of
  $18{,}790$ images assembled from eight public sources, deduplicated and cleaned
  to span diverse indoor scenes, and partitioned for federated (IID and non-IID)
  evaluation (Sec.~\ref{sec:system}).
  \item \textbf{(C2) Edge-deployable federated detector.} A fire detector built
  on a frozen MobileNet-V2 feature extractor with a lightweight trainable head,
  whose head updates are compressed to reduce per-round uplink up to $10\times$
  via 8-bit and 4-bit integer (INT8/INT4) quantization and error-feedback
  Top-$K$ sparsification (Secs.~\ref{sec:system}--\ref{sec:comm}).
  \item \textbf{(C3) Semi-decentralized Byzantine-robust FL method.} A
  history-aware aggregation rule (SafeguardSGD), composable with the compressors
  of (C2), run on a \emph{rotating coordinator} that preserves each client's
  detection state across handoffs---removing the fixed-server single point of
  failure while matching fixed-server balanced accuracy within $0.001$ in our
  experiments, and evicting stealthy low-magnitude attacks that per-round
  filters (Krum, centered clipping) miss (Secs.~\ref{sec:byz}--\ref{sec:rotate}).
\end{itemize}

\section{System and Data}\label{sec:system}

\noindent\textbf{Dataset.} We assemble a binary (fire / no-fire) indoor
image dataset of $18{,}790$ images from eight public sources, deduplicated and
cleaned, spanning diverse indoor scenes (homes, offices, warehouses,
corridors). Exploratory analysis shows the two classes are separated primarily
by \emph{localized} high-intensity regions (flames) rather than global
brightness, a property later confirmed by explainability analysis. Of the
$18{,}790$ images, $9{,}206$ form a class-balanced training split, $5{,}152$ a
validation split, and $4{,}432$ a held-out test split. Training and validation
data are partitioned across five clients under both an IID split and a non-IID
label-skew split ($\approx\!70\%$ majority class per client). Each client uses
its local validation split for early stopping and model selection, while the
global test set, used only for final evaluation, is kept fixed. Full curation
and partitioning details are documented in the underlying
thesis~\cite{argyrou2026}.

\noindent\textbf{Model.} Each client uses an ImageNet-pretrained
MobileNet-V2~\cite{sandler2018} backbone, chosen for its favorable
accuracy/compute balance on constrained devices. The backbone is \emph{frozen} and its $1280$-dimensional global-average-pooled
feature vector is fed to a two-layer head ($1280\!\to\!256\!\to\!2$ with ReLU and
dropout). This design has two benefits for edge FL\@. Features are pre-extracted
once and cached on disk, so each federated round reduces to a lightweight head
optimization rather than a full backbone pass. Moreover, only the head's $N=328{,}450$
parameters are exchanged, shrinking the communication payload. Training uses a
class-weighted cross-entropy loss ($w_{\text{fire}}=2$) reflecting the
asymmetric cost of a missed fire, and metrics are reported at a fire-biased
operating point of $0.4$. Because accuracy can be misleading under class
imbalance, we report \emph{balanced accuracy} (BA) and fire recall as primary
metrics, averaged over three seeds. Clients optimize the head with Adam
($10^{-3}$), batch size $16$, two local epochs per round for up to $50$ rounds.
SafeguardSGD's window sizes and threshold follow the values recommended
in~\cite{allenzhu2021}, chosen there to balance rapid detection of persistent
attackers against false eviction of honest clients; they were
not tuned on the test set.

\noindent\textbf{Federated baseline.} With honest clients, FedAvg on this
system reaches test BA $0.9835\!\pm\!0.0014$ (IID) and
$0.9818\!\pm\!0.0026$ (non-IID) with fire recall $\approx\!0.98$, above the
centralized baseline (BA $0.9615$). Heterogeneity-aware
aggregators (FedProx, FedAdam) did not surpass FedAvg under the mild label skew
studied, so FedAvg is used as the honest reference below.

\section{Communication-Efficient Aggregation}\label{sec:comm}
On bandwidth-limited edge networks the per-round uplink can be a practical
bottleneck. We apply two message-level compressors that are agnostic to the
aggregation rule and can be composed with any Byzantine-robust strategy.
\emph{Uniform quantization}~\cite{alistarh2017} maps each float32 head parameter to a $b$-bit
integer: 8-bit (INT8; $1$\,B/element, $4\times$) and 4-bit (INT4; bit-packed,
$8\times$), with per-tensor min-max scaling and no error feedback. \emph{Top-$K$
sparsification} transmits only the $k$ fraction of largest-magnitude update
elements as (value, index) pairs and accumulates the discarded residual locally
via error feedback~\cite{stich2018}, so information is reintroduced over
successive rounds.

\begin{figure}[t]
\centering
\includegraphics[width=0.86\linewidth]{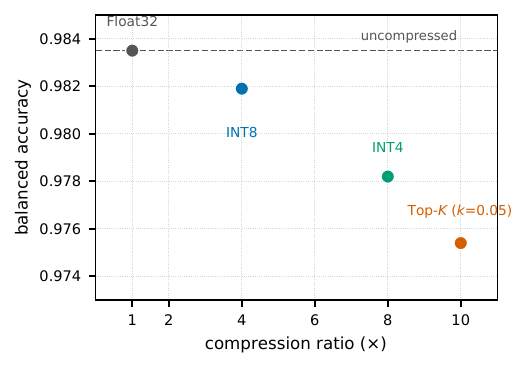}
\caption{Communication trade-off (IID, MobileNet-V2). Global test balanced
accuracy (vertical axis) versus per-client uplink compression ratio (horizontal
axis, log-spaced) for the four schemes of Table~\ref{tab:comm}. Accuracy falls
by only ${\approx}0.8$ points (from $0.9835$ uncompressed to $0.9754$ at
$10\times$), a gradual degradation.}\label{fig:comm}
\end{figure}

Fig.~\ref{fig:comm} and Table~\ref{tab:comm} report the trade-off. INT8
($4\times$) reduces BA by only $0.0016$ (from $0.9835$ to $0.9819$); INT4
($8\times$) and error-feedback Top-$K$ at $k\!=\!0.05$ ($10\times$) cost
$0.5$--$0.8$ BA points (down to $0.9782$ and $0.9754$) for higher compression.
The $328{,}450$-parameter head shrinks from a $1283$\,KB float32 payload to
$128$\,KB\@. Encoding overhead is negligible relative to a
$\sim\!30$\,s local training step ($\le\!17$\,ms per round).

\begin{table}[t]
\centering
\caption{Communication trade-off (IID, MobileNet-V2). ``Payload'' is the
per-client, per-round wire size in kilobytes; ``Ratio'' is the compression
factor over the float32 baseline; ``BA'' is global test balanced accuracy.}\label{tab:comm}
\small
\begin{tabular}{@{}lccc@{}}
\toprule
Scheme & Payload (KB) & Ratio & BA \\
\midrule
Float32 (baseline)      & 1283.0 & $1.0\times$  & 0.9835 \\
INT8 quantization       & 320.8  & $4.0\times$  & 0.9819 \\
INT4 quantization       & 160.4  & $8.0\times$  & 0.9782 \\
Top-$K$ ($k\!=\!0.05$)  & 128.3  & $10.0\times$ & 0.9754 \\
\bottomrule
\end{tabular}
\end{table}

\section{History-Aware Byzantine Robustness}\label{sec:byz}
We consider a synchronous cross-device setting with $n=5$ clients and at most
$f=1$ Byzantine client (fraction $0.2$), a non-omniscient adversary that
controls a single compromised camera and cannot observe honest clients'
gradients. The adversary is \emph{static}: the same client remains malicious
throughout all rounds, the setting under which permanent eviction from the
retained set is the appropriate response. We study five attacks spanning \emph{overt} behaviors that cause
immediate divergence and \emph{covert} behaviors that inject small persistent
biases: sign-flip ($g_i\!\leftarrow\!-g_i$), scaling ($g_i\!\leftarrow\!\gamma
g_i$), additive Gaussian noise, \emph{stealth-flip}
($g_i\!\leftarrow\!(1-\alpha)g_i$, a partial inversion with $\alpha\in[0,1]$),
and inner-product manipulation (IPM, $g_i\!\leftarrow\!-c\,g_i$ with small
$c$). The last two keep the submitted vector geometrically close to the honest
cluster while contributing a persistent anti-gradient component each round.

\noindent\textbf{Baseline defenses.} \emph{Krum}~\cite{blanchard2017} is a
stateless per-round filter that selects the update with the smallest sum of
distances to its nearest neighbors. \emph{Centered clipping (CClip)}~\cite{karimireddy2021} maintains a
server momentum vector and clips updates beyond a radius $\tau$. Both are blind
to attacks that stay geometrically central or below the clipping radius.

\noindent\textbf{SafeguardSGD.} SafeguardSGD~\cite{allenzhu2021} was proposed
for Byzantine-resilient distributed SGD with a fixed server; it augments
federated SGD with a dual-window anomaly detector; carrying its per-client
state across coordinator handoffs (Sec.~\ref{sec:rotate}) is what sets our
approach apart from prior rotating-server schemes. For each client the
coordinator maintains two running sums of that client's submitted updates,
over a short window ($T_0\!=\!10$ rounds) and a long window ($T_1\!=\!30$
rounds), and measures how far each running sum has drifted from the
honest-cluster (geometric-median) direction. A \emph{persistent}
Byzantine client's deviation grows linearly in the number of rounds and
eventually crosses the threshold $\tau\!=\!0.75$, whereas an honest client's
bounded fluctuation does not. Flagged clients are permanently removed from the
retained set $\mathcal{G}_t$ (SafeguardSGD's ``good worker'' set); a norm floor $\eta_{\text{floor}}\!=\!0.5$
prevents premature eviction during warm-up. Crucially, this history-based
criterion can detect stealthy attacks whose \emph{individual-round} deviation is
too small to trip a geometric threshold but whose effect accumulates.

\noindent\textbf{Results.} Besides test BA we report the \emph{eviction
round}---the round at which the attacker is permanently removed, averaged over
seeds. It measures the attacker's exposure window: every earlier round lets
the corrupted update enter the aggregate, so lower is better.
Table~\ref{tab:byz} compares the three defenses. Under overt attacks all
three preserve performance, and SafeguardSGD evicts the attacker within
$1$--$3$ rounds. The defenses diverge on stealthy attacks. Krum's per-round
geometry leaves it most exposed in the intermediate stealth-flip regime
($\alpha\!\approx\!0.4$--$0.5$), where the malicious gradient is still selected
in $\sim\!35\%$ of rounds; CClip provides no defense once the attenuated
gradient falls inside $\tau$ ($0$ malicious clips for $\alpha\!\ge\!0.5$).
SafeguardSGD closes this gap by exploiting its long-window accumulator: it
detects and evicts IPM at round $6.3\!\pm\!0.5$ and stealth-flip for
$\alpha\!\ge\!0.3$, after which metrics recover to within noise of the no-attack
baseline. The only regime where it offers no improvement is very-weak
stealth-flip ($\alpha\!\le\!0.2$), where the drift is statistically
indistinguishable from honest variance---but there the attack is also too weak
to cause meaningful harm ($\ba$ within $0.001$ of baseline). Under no attack
SafeguardSGD attains BA $0.9575$, within $0.004$ of Krum ($0.9613$) on the same
pipeline.

\begin{figure}[t]
\centering
\includegraphics[width=0.94\linewidth]{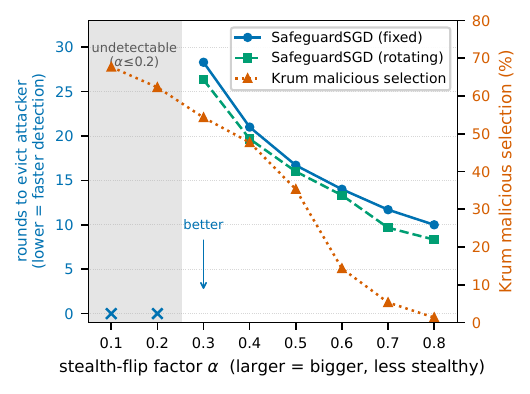}
\caption{Byzantine robustness under the stealth-flip attack, in which the
malicious client submits $(1\!-\!\alpha)$ times its true gradient (IID,
MobileNet-V2). A larger $\alpha$ is a bigger but \emph{less stealthy}
perturbation: the stealthiest attacks sit at small $\alpha$ and the most harmful
around $\alpha\!\approx\!0.4$--$0.5$. \emph{Left axis (blue):} the round at which
SafeguardSGD permanently evicts the attacker---\emph{lower is faster, hence
better}---for the fixed (circles) and rotating (squares) architectures, which
overlap; a ``$\times$'' on the baseline marks runs where the attacker was never
evicted within the $100$-round budget. \emph{Right axis (orange):} the fraction
of rounds in which Krum selects the malicious client's update; unlike
SafeguardSGD, Krum never permanently evicts it. The shaded band
($\alpha\!\le\!0.2$) is too weak to detect but also too weak to harm accuracy.
Takeaway: SafeguardSGD detects and removes the attacker---faster as $\alpha$
grows---while Krum keeps trusting it.}\label{fig:byz}
\end{figure}

Fig.~\ref{fig:byz} makes the mechanism explicit over the full stealth-flip
sweep. SafeguardSGD's history accumulator evicts the attacker for
$\alpha\!\ge\!0.3$, and sooner as $\alpha$ grows (the eviction round falls from
$28$ at $\alpha\!=\!0.3$ to $10$ at $\alpha\!=\!0.8$), whereas Krum's per-round
filter never evicts and keeps selecting the malicious update in a substantial
fraction of rounds throughout the harmful intermediate regime. Notably, the
fixed-server and rotating variants (Sec.~\ref{sec:rotate}) evict at the same
rounds up to seed variability (Table~\ref{tab:rotate}).

\begin{table}[t]
\centering
\caption{Byzantine robustness (IID, MobileNet-V2, malicious client 3): test
balanced accuracy over three seeds. ``Evict'' is SafeguardSGD's mean eviction
round (lower is better; --~=~not evicted); it does not apply to Krum and
CClip, which filter per round but never evict.}\label{tab:byz}
\small
\setlength{\tabcolsep}{4pt}
\begin{tabular}{@{}lcccc@{}}
\toprule
Attack & Krum & CClip & SafeguardSGD & Evict \\
\midrule
No attack               & 0.9613 & --     & 0.9575 & --   \\ 
Sign-flip               & 0.9601 & 0.9501 & 0.9553 & 3.0  \\
Scaling                 & 0.9609 & 0.9667 & 0.9571 & 2.0  \\
Noise                   & 0.9610 & 0.9564 & 0.9546 & 1.0  \\
Stealth-flip ($\alpha\!=\!0.5$) & 0.9543 & 0.9528 & 0.9550 & 16.7 \\
IPM ($c\!=\!0.1$)       & 0.9594 & 0.9539 & 0.9545 & 6.3  \\
\bottomrule
\end{tabular}
\end{table}

\section{Rotating-Coordinator Architecture}\label{sec:rotate}
The defenses above still assume a fixed coordinator (the node that performs
aggregation): if the coordinator becomes unavailable the federation stalls, and
trust is concentrated in a single node. We remove this single point of failure
by \emph{rotating the coordinator}, reassigning the role to a different edge
node each round.
Among the aggregators studied, SafeguardSGD carries the most per-client state,
making it the informative case for rotation: at each handoff the outgoing
coordinator serializes the global weights, the per-client accumulators
$A_i,B_i$, the retained set $\mathcal{G}_t$ and participation history, and the
elected next coordinator's identifier, so that detection state survives the role
transfer.

\noindent\textbf{Eligibility and election.} At the end of each round a client is
eligible for the next coordinator role iff it (i) belongs to the current
retained set $\mathcal{G}_t$ (never flagged Byzantine), (ii) meets an availability threshold
($\ge\!p_{\min}\!=\!0.8$ of the last $w_a\!=\!5$ rounds), and (iii) responded in
\emph{all} of the last $w_f\!=\!3$ rounds. Among eligible candidates the next
coordinator is chosen by deterministic round-robin over partition identifiers,
requiring no centralized scheduler. We physically realize rotation by starting a
fresh Flower \texttt{SuperLink} on the elected node's Amazon Web Services (AWS)
Elastic Compute Cloud (EC2) instance each round.

\noindent\textbf{Results.} Table~\ref{tab:rotate} compares fixed-server and
rotating SafeguardSGD\@. The rotating variant matches detection speed and final
accuracy across the full attack sweep: no-attack BA $0.9565\!\pm\!0.0006$ is
within $0.001$ of the fixed-server value ($0.9575$), and stealth-flip/IPM
eviction rounds agree ($16.0$ vs.\ $16.7$ and $6.3$ vs.\ $6.3$). In the
no-attack case the round-robin
distributed leadership \emph{perfectly uniformly} across all five nodes
($20/20/20/20/20$ coordinator rounds), empirically confirming the eligibility
parameters are well-calibrated. The one structural difference from the
fixed-server setting is that, before its eviction, a malicious client can itself
be elected coordinator and briefly control aggregation (e.g.,\ once, for IPM).
Empirically this coordinator-exposure window did not degrade final model
quality (IPM BA $0.9545$ under both architectures), but a fixed server never
hands the attacker this role, so safety-critical deployments should keep this
risk in mind.

\begin{table}[t]
\centering
\caption{Fixed vs.\ rotating SafeguardSGD (IID, MobileNet-V2). Each row
reports test balanced accuracy (BA) and SafeguardSGD's mean eviction round
(``Evict'') for one attack.}\label{tab:rotate}
\small
\setlength{\tabcolsep}{4pt}
\begin{tabular}{@{}lcccc@{}}
\toprule
 & \multicolumn{2}{c}{Fixed server} & \multicolumn{2}{c}{Rotating} \\
\cmidrule(lr){2-3}\cmidrule(lr){4-5}
Attack & BA & Evict & BA & Evict \\
\midrule
No attack             & 0.9575 & --   & 0.9565 & --   \\ 
Sign-flip             & 0.9553 & 3.0  & 0.9575 & 1.0  \\
Scaling               & 0.9571 & 2.0  & 0.9605 & 1.0  \\
Noise                 & 0.9546 & 1.0  & 0.9537 & 1.0  \\
Stealth-flip ($0.5$)  & 0.9550 & 16.7 & 0.9578 & 16.0 \\
IPM ($c\!=\!0.1$)     & 0.9545 & 6.3  & 0.9545 & 6.3  \\
\bottomrule
\end{tabular}
\end{table}

\noindent\textbf{Deployment.} The system was validated on six distributed EC2
instances within a single virtual private cloud: one \texttt{SuperLink}
coordinator (rotated) and five \texttt{SuperNode} clients communicating over
gRPC, with cached per-client features and non-overlapping partitions. This
confirms the mechanisms transfer from simulation to a physically distributed
edge-computing environment. All images originate from public sources; full
curation, split, hyperparameter, and deployment details appear
in~\cite{argyrou2026}.

\noindent\textbf{Explainability.} Gradient-weighted class activation mapping
(Grad-CAM) and occlusion-sensitivity analysis of
the final model trained under the rotating coordinator confirm that predictions are driven by
localized, high-intensity flame regions rather than spurious background
correlations, consistent with the localized-brightness signal identified in the
data analysis.

\section{Conclusion}\label{sec:conclusion}
We presented a federated indoor fire-detection system that simultaneously
addresses communication cost, Byzantine robustness, and the fixed-server single
point of failure. History-aware SafeguardSGD aggregation detects and permanently
evicts stealthy low-magnitude attacks that per-round geometric filters miss,
and a rotating-coordinator architecture preserves this behavior while
distributing the aggregation role uniformly across edge nodes.
On an $18{,}790$-image dataset and a distributed AWS~EC2 deployment the system
matches the centralized baseline's balanced accuracy, reduces uplink up to
$10\times$, and tolerates a Byzantine client without a fixed trusted server.
Our evidence is empirical, obtained with five clients and a single static,
non-colluding adversary. Future work includes adaptive, multi-client and
colluding threat models, verifiable aggregation, and larger-scale deployments.

\vfill\pagebreak

\bibliographystyle{IEEEbib}

\end{document}